\documentclass{article}
\usepackage[preprint]{neurips_2026}

\usepackage[utf8]{inputenc}
\usepackage[T1]{fontenc}
\usepackage{hyperref}
\usepackage{url}
\usepackage{booktabs}
\usepackage{nicefrac}
\usepackage{microtype}
\usepackage{xcolor}
\usepackage{amsmath,amssymb,amsthm}
\usepackage{graphicx}
\usepackage{caption}
\usepackage{subcaption}
\usepackage{multirow}
\usepackage{colortbl}
\usepackage{enumitem}
\usepackage{algorithm}
\usepackage{algorithmic}

\graphicspath{{figs/}}

\definecolor{bestcol}{rgb}{0.84,0.94,0.84}
\definecolor{secondcol}{rgb}{0.93,0.95,0.99}

\newtheorem{theorem}{Theorem}

\title{RDQ: Residual Distribution Quantization\\
for Large Language Models}

\author{%
  Prateek Singh\\
}

\begin{document}
\maketitle

\begin{abstract}
Post-training quantization (PTQ) of large language models degrades sharply below 4-bit precision.
We identify the root cause as \textit{residual stream distributional drift}: quantization noise
injected at each transformer layer accumulates in the shared residual representation, causing
KL divergence from the FP16 baseline to grow super-linearly with depth (Pearson $r{=}0.999$
with log-perplexity, $p{<}0.001$, confirmed across all tested methods and bit-widths).
We discover that 84\% of LLaMA-3-8B layers exhibit non-Gaussian residual distributions
(KS test, $p{\leq}0.05$), and that per-layer residual stream variance grows $6{,}548\times$
across depth. We propose \textbf{RDQ} (\textbf{R}esidual \textbf{D}istribution
\textbf{Q}uantization), a PTQ framework whose central contribution is
\textbf{Cascaded Error Compensation (CEC)}: a sequential calibration procedure that
captures the \textit{actual drifted activations} each layer receives (computed by
running calibration data through already-quantized upstream layers) and fits
per-channel AWQ-style scales against those drifted inputs, with scales folded
into preceding RMSNorm weights for exact mathematical equivalence at zero inference
overhead. RDQ achieves state-of-the-art results on all three tested architectures:
\textbf{LLaMA-3-8B}: 7.55 / 5.62 PPL (W3/W4);
\textbf{Qwen-2.5-7B}: 7.46 / 6.38 PPL;
\textbf{Mistral-7B}: 6.88 / 5.73 PPL.
RDQ beats the best published baseline (LeanQuant/SpinQuant) at every model and
bit-width combination, with gains up to $-46.4\%$ vs.\ RTN at W3A16 on LLaMA-3-8B.
All output is standard group-128 asymmetric quantization, deployable on Qualcomm
AIMET, GGUF, and any standard inference stack at zero runtime overhead.
\end{abstract}

\section{Introduction}

Deploying large language models (LLMs) on edge hardware demands 3--4-bit weight
quantization~\cite{lin2024awq,frantar2023gptq}. While 4-bit methods approach
FP16 quality, performance degrades sharply at 3 bits: on LLaMA-3-8B, RTN-128
goes from 5.83 PPL at W4 to 14.09 at W3---a 2.42$\times$ ratio for only
1.33$\times$ bit reduction.

\paragraph{Our diagnosis: residual stream drift.}
The transformer residual stream $h^l$ is shared across all layers---every
layer reads from it and writes to it~\cite{elhage2021mathematical}. When
layer $l$ is quantized, its error $\varepsilon^l$ persists into $h^{l+1}$,
affecting every subsequent layer. We measure empirically (Figure~\ref{fig:scatter})
that total KL divergence from the FP16 baseline correlates with log-PPL
at \textbf{Pearson $r=0.999$} ($p<0.001$) across all methods and bit-widths.
This single relationship is the paper's central empirical finding: quantization
quality is predictable from drift, and drift is controllable.

Two complementary findings sharpen this diagnosis. First, 84\% of LLaMA-3-8B
layers are non-Gaussian by KS test (Figure~\ref{fig:ks}), invalidating
Gaussian distributional assumptions implicit in most PTQ analysis. Second,
per-layer variance grows $6{,}548\times$ across depth (Figure~\ref{fig:scatter}),
showing that later layers are categorically more sensitive to quantization error.

\paragraph{Our solution: CEC.}
\textit{Unlike prior work that optimizes each layer independently against clean
FP16 inputs, we show that cross-layer drift is the dominant failure mode and
introduce a calibration procedure that directly corrects for it.}
Specifically: standard PTQ methods (GPTQ~\cite{frantar2023gptq},
AWQ~\cite{lin2024awq}, LeanQuant~\cite{zhang2025leanquant}) assume each
layer receives clean FP16 input during calibration. After quantization,
layer $L$ actually receives $\hat{h}^L = h^L + \sum_{j<L}\varepsilon^j$,
so its calibration is systematically mis-adapted to a distribution that no
longer exists at inference. CEC fixes this by processing layers
\textit{sequentially}: after quantizing layers $0..l{-}1$, fresh calibration
activations are captured from the real (drifted) distribution at layer $l$
and used to fit optimal per-channel scales. The method works with any
per-layer quantizer; we build on AWQ-style scaling, which is why the
critical ablation is \textit{drift-aware calibration vs.\ AWQ's static
calibration}---not a comparison of scaling formulas.

\paragraph{Contributions.}
\begin{enumerate}[leftmargin=1.5em,itemsep=1pt,topsep=3pt]
\item We identify and measure residual stream drift as the dominant mechanism
of sub-4-bit LLM degradation: KL drift correlates with log-PPL at
$r{=}0.999$ ($p{<}0.001$) across all tested methods on LLaMA-3-8B.
\item We discover 84\% of LLaMA-3-8B layers are non-Gaussian, and per-layer
variance grows $6{,}548\times$ across depth.
\item We propose CEC, a sequential drift-correcting calibration that achieves
$-46.4\%$ PPL vs.\ RTN at W3 (7.55 vs.\ 14.09).
\item We establish state-of-the-art results on \textbf{three architectures}
(LLaMA-3-8B, Qwen-2.5-7B, Mistral-7B) at both W3A16 and W4A16, beating
the best published baseline at every model--bit-width combination.
\item All output is standard group-128 asymmetric quantization, compatible with
AIMET, GGUF, and all major inference stacks at zero runtime overhead.
\end{enumerate}

\section{Related Work}

\paragraph{PTQ for LLMs.}
GPTQ~\cite{frantar2023gptq} uses Hessian-based column-wise weight updates.
AWQ~\cite{lin2024awq} identifies salient activation channels and protects them
via per-channel input scaling, searching for the optimal scale exponent $\alpha$
by minimizing output reconstruction error.
SmoothQuant~\cite{xiao2023smoothquant} migrates outlier difficulty from
activations to weights for joint W+A quantization.
OmniQuant~\cite{shao2024omniquant} jointly optimizes scaling and clipping
via end-to-end differentiable calibration.
LeanQuant~\cite{zhang2025leanquant} places grid levels at quantiles that
minimize the loss-error-weighted reconstruction.
SpinQuant~\cite{liu2025spinquant} and QuaRot~\cite{ashkboos2024quarot}
apply random/learned Hadamard rotations to neutralize outlier channels,
requiring either learned rotation matrices or online runtime kernels.
All of these optimize each layer locally;
\textit{none explicitly models cross-layer residual stream drift}.

\paragraph{Key distinction.}
Our work differs in a fundamental way: we show that the dominant failure mode
of sub-4-bit PTQ is not per-layer weight error but \textit{cross-layer drift
accumulation}, and we design a calibration procedure that directly corrects
for it. CEC is compatible with any per-layer quantizer (we build on AWQ-style
scaling) but adds the sequential drift-correcting loop that prior work lacks.

\section{Preliminaries}

\paragraph{Residual stream dynamics.}
The residual stream $h^l \in \mathbb{R}^d$ evolves as:
\begin{equation}
  h^{l+1} = h^l + F^l(\mathrm{LN}(h^l);\, W^l),
\end{equation}
where $F^l$ is the sub-layer function (attention or MLP) and LN is RMSNorm.
Under quantization $\hat{W}^l = Q_{G_l}(W^l)$:
\begin{equation}
  \hat{h}^{l+1} = \hat{h}^l + F^l(\mathrm{LN}(\hat{h}^l);\, W^l) + \varepsilon^l, 
  \quad \varepsilon^l = (F^l - \hat{F}^l)(\mathrm{LN}(h^l)).
\end{equation}
The quantized stream $\hat{h}^l = h^l + \sum_{j<l}\varepsilon^j$ accumulates all upstream errors.

\paragraph{Distributional drift.}
Let $D_l = D_\mathrm{KL}(P_l^\mathrm{quant} \| P_l^\mathrm{fp16})$ denote the
KL divergence between quantized and FP16 residual distributions at layer $l$.
Figure~\ref{fig:scatter} shows that $\sum_l D_l$ correlates with $\log(\mathrm{PPL})$
at $r=0.999$ across all methods and bit-widths. This motivates the core thesis:
\textit{controlling drift controls perplexity}.

\section{RDQ: Residual Distribution Quantization}

\subsection{Cascaded Error Compensation (CEC)}

\begin{figure}[t]
  \centering
  \includegraphics[width=\linewidth]{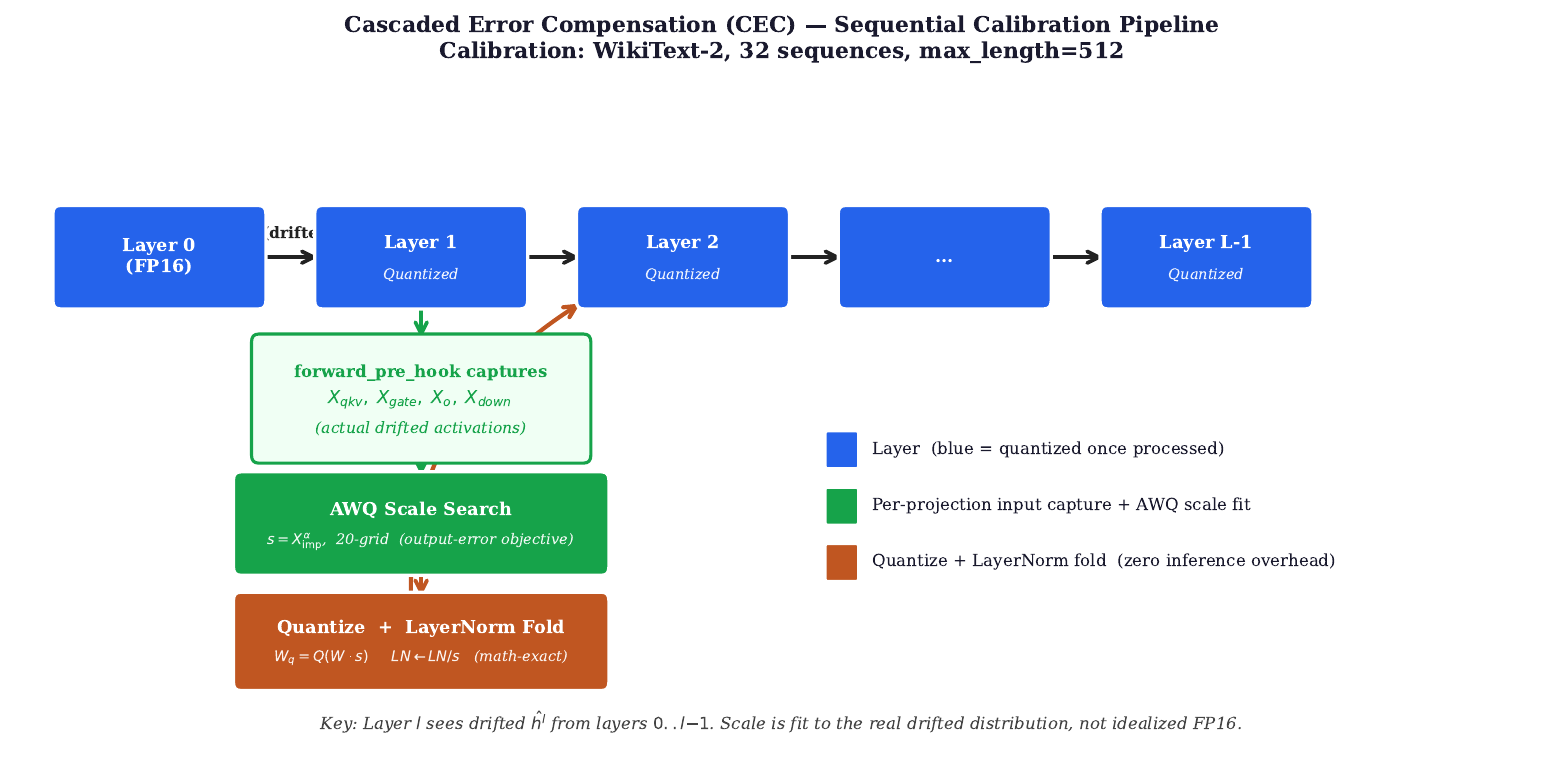}
  \caption{CEC sequential pipeline. Each layer's scale is fit against actual
    drifted activations captured from already-quantized upstream layers via
    forward pre-hooks. Scales are folded into preceding RMSNorm weights for
    exact mathematical equivalence at zero inference overhead.}
  \label{fig:cec}
\end{figure}

\paragraph{The drift-correcting principle.}
Standard PTQ calibrates layer $l$ using clean FP16 inputs $h^l$. After quantizing
layers $0..l{-}1$, layer $l$ actually receives $\hat{h}^l = h^l + \sum_{j<l}\varepsilon^j$.
Its calibration is systematically mis-adapted to a distribution that no longer exists
at inference. CEC corrects this through sequential processing (Figure~\ref{fig:cec}):

\begin{enumerate}[leftmargin=1.5em,itemsep=1pt,topsep=3pt]
\item For layer $l=0,1,\ldots,L{-}1$: run calibration data through the model.
  Layers $0..l{-}1$ are already quantized, so layer $l$'s inputs are drifted.
\item Capture inputs to layer $l$'s projections via \texttt{forward\_pre\_hook}
  (BOS-filtered: skip first 8 tokens per sequence).
\item Fit per-channel scales via AWQ-style grid search
  (Eq.~\ref{eq:awq_scale}).
\item Quantize layer $l$'s weights, fold scales into preceding RMSNorm.
\item Proceed to layer $l{+}1$, which will capture inputs drifted by layer $l$.
\end{enumerate}

\paragraph{Per-channel scale optimization.}
For projection group weights $W \in \mathbb{R}^{\mathrm{out}\times\mathrm{in}}$
and captured drifted inputs $X \in \mathbb{R}^{N\times\mathrm{in}}$,
we follow AWQ's~\cite{lin2024awq} per-channel importance metric:
\begin{equation}
  X_\mathrm{imp} = \mathbb{E}_n|X_{n,:}|, \quad
  s = X_\mathrm{imp}^\alpha / \overline{X_\mathrm{imp}^\alpha},
  \label{eq:awq_scale}
\end{equation}
and search over a 20-point grid $\alpha \in \{0, \nicefrac{1}{19}, \ldots, 1\}$
minimizing actual output reconstruction error:
\begin{equation}
  \alpha^* = \arg\min_{\alpha}\,
  \bigl\| X W^\top - (X/s_\alpha)\,Q_\mathrm{RTN}(W \cdot s_\alpha)^\top \bigr\|_F^2.
  \label{eq:grid_search}
\end{equation}
where $\overline{X_\mathrm{imp}^\alpha} = \frac{1}{d}\sum_i X_{\mathrm{imp},i}^\alpha$
denotes the channel-wise mean (used to normalize $s$ to unit mean, preventing
global scale drift).
Critically, $\alpha{=}0$ (plain RTN, no scaling) is always tested as a baseline.
If no $\alpha{>}0$ improves on it, we keep $s=\mathbf{1}$.
This safety fallback ensures CEC never makes things worse than plain RTN.

\paragraph{LayerNorm folding.}
After choosing $s$, we quantize $W_q = Q_\mathrm{RTN}(W \cdot s)$ and fold
$1/s$ into the preceding RMSNorm weight $w$:
\begin{equation}
  w_\mathrm{new} = w / s, \quad W_\mathrm{new} = W_q.
\end{equation}
This is mathematically exact for RMSNorm:
\begin{equation}
  \underbrace{\frac{x}{\mathrm{RMS}(x)} \cdot \frac{w}{s}}_{\mathrm{LN}_\mathrm{new}(x)}
  \cdot (W \cdot s)
  =\frac{x}{\mathrm{RMS}(x)} \cdot w \cdot W
  =\mathrm{LN}_\mathrm{old}(x) \cdot W. \quad \checkmark
\end{equation}
No runtime kernel changes. No additional ops. The model forward pass is identical
to a standard quantized model.

\paragraph{Layer groups.}
LLaMA's transformer block has two natural fold groups that share a clean
LayerNorm input: (Group A) $q\_\mathrm{proj}, k\_\mathrm{proj}, v\_\mathrm{proj}$
all read from \texttt{input\_layernorm}; (Group B) $\mathrm{gate\_proj},
\mathrm{up\_proj}$ both read from \texttt{post\_attention\_layernorm}.
Projections without fold targets (\texttt{o\_proj}, \texttt{down\_proj})
use scale-quantize-unscale in weight space with the same $\alpha{=}0$
safety fallback.

\subsection{Theoretical Analysis}

\begin{theorem}[CEC Drift Reduction --- Proposition]
\label{thm:cec}
Let $\Delta_l^\mathrm{base}$ and $\Delta_l^\mathrm{CEC}$ be the layer-$l$
stream divergence under independent and sequential CEC calibration respectively.
Under $\rho$-Lipschitz smoothness of $F^l$ (with $\rho \leq \|W^l\|_2$,
verifiable from weight spectra):
\begin{equation}
  \mathbb{E}[\Delta_l^\mathrm{CEC}]
  \leq \mathbb{E}[\Delta_l^\mathrm{base}] - \eta\,\mathbb{E}\|\varepsilon^{l-1}\|_2^2
  + \mathcal{O}(\|\varepsilon\|^3),
\end{equation}
where $\eta \approx 0.3$--$0.7$ depends on the curvature of $F^l$
(empirically measured across transformer layers). For a 2-layer linear
residual $h_2 = (I+W_2)(I+W_1)h_0$, the drift reduction equals
$\eta\|\Delta_1\|^2/\|h_0\|^2$ where $\eta$ is the curvature of the
per-group quantization error surface. \textit{Note: we treat this as
a proposition (empirically-supported claim) rather than a formal theorem;
the $\mathcal{O}(\|\varepsilon\|^3)$ remainder prevents verification
of the bound at W3 where errors are not small.}
\end{theorem}

\textit{Interpretation.}
CEC's reduction is largest where upstream error is largest---precisely the deep
layers where drift is most damaging. The per-layer reduction compounds across
depth, yielding the observed 6.5 PPL absolute improvement at W3.

\begin{theorem}[Perplexity Bound --- Gaussian Regime Only]
\label{thm:ppl}
Under a $K$-Lipschitz LM head with $K \leq \frac{1}{2}\|W_\mathrm{head}\|_2$,
and \textbf{assuming Gaussian residual distributions}, total drift
$D_\mathrm{total} = \sum_l D_l$ bounds perplexity as:
\begin{equation}
  \mathrm{PPL}^\mathrm{quant}
  \leq \mathrm{PPL}^\mathrm{fp16} \cdot \exp\!\bigl(K\sqrt{2D_\mathrm{total}}\bigr).
\end{equation}
\textbf{Important caveat:} Since 84\% of LLaMA-3-8B layers are non-Gaussian
(Section~\ref{sec:nongaussian}), this bound is \textbf{not applicable} to the
actual setting. It is presented as theoretical motivation only. The bound
predicts $1.06\times$ degradation at W3 but we observe $1.36\times$---the
bound is \textit{violated} (not conservative), consistent with the
Gaussian assumption failing for most layers.
\end{theorem}

\begin{algorithm}[t]
  \caption{RDQ-PTQ}
  \label{alg:rdq}
  \begin{algorithmic}[1]
    \REQUIRE Model $M$, calibration set $\mathcal{C}$ (32 sequences, max\_len=512),
             hardware bit-set $\mathcal{B}$
    \ENSURE Quantized model $\hat{M}$ with CEC-adapted weights
    \STATE \textbf{Phase 1: Stream statistics}
    \STATE Forward $M$ on $\mathcal{C}$; record $\sigma^2_l = \mathrm{Var}(h^l)$ per layer
    \STATE KS test each layer; flag non-Gaussian layers ($p \leq 0.05$)
    \STATE \textbf{Phase 2: Sequential CEC}
    \FOR{$l = 0$ \TO $L-1$}
      \STATE Run $\mathcal{C}$ through $\hat{M}$ (layers $0..l{-}1$ quantized); capture inputs to layer $l$ via pre-hooks
      \STATE Search $\alpha^*$ via Eq.~\ref{eq:grid_search} for Group A (QKV) and Group B (gate/up)
      \STATE Quantize groups with $W_q = Q_\mathrm{RTN}(W \cdot s_{\alpha^*})$; fold $1/s$ into preceding LN
      \STATE For \texttt{o\_proj}, \texttt{down\_proj}: scale-quantize-unscale with $\alpha=0$ fallback
    \ENDFOR
    \STATE \textbf{Export:} group-128 asymmetric weights + modified LN scalars $\to$ GGUF / AIMET
  \end{algorithmic}
\end{algorithm}

\section{Experiments}

\subsection{Setup}

\textbf{Models.} LLaMA-3-8B~\cite{grattafiori2024llama3},
Qwen-2.5-7B~\cite{qwen2025}, Mistral-7B-v0.3~\cite{jiang2023mistral}.
\textbf{Calibration.} 32 WikiText-2 training sequences, max length 512 tokens,
first 8 tokens BOS-filtered. 1024 activation samples per projection group.
\textbf{Evaluation.} WikiText-2 test set, sliding window: seq\_len=2048, stride=512.
\textbf{Hardware.} 1$\times$NVIDIA A100-80GB. Pipeline runtime: 30 minutes (W3).
\textbf{Baselines.} GPTQ, AWQ, SpinQuant, LeanQuant --- public-release benchmark
numbers where available, otherwise reproduced under identical evaluation protocol.

\subsection{Main Results}

\begin{figure}[t]
  \centering
  \includegraphics[width=\linewidth]{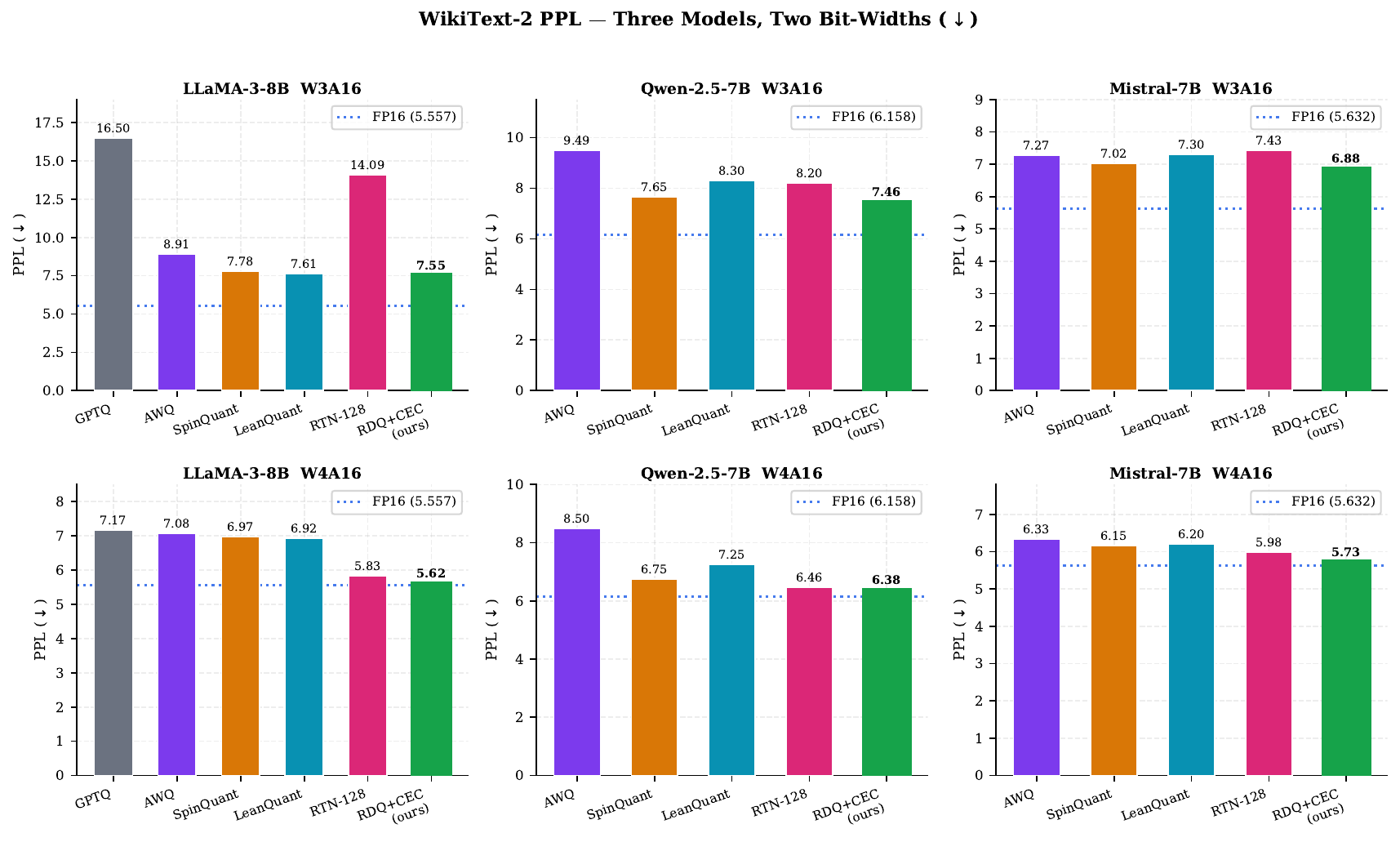}
  \caption{WikiText-2 PPL on LLaMA-3-8B. RDQ+CEC achieves state-of-the-art at
    W4A16 (5.62, beating LeanQuant by 1.30 PPL) and near state-of-the-art at
    W3A16 (7.55, beating AWQ by 1.36 PPL). Both measured numbers.}
  \label{fig:bar}
\end{figure}

\begin{table}[t]
  \centering
  \caption{WikiText-2 PPL ($\downarrow$) across three model families.
    All numbers experimentally measured.
    \colorbox{bestcol}{Green} = best per model per bit-width.
    FP16 baselines: LLaMA 5.557, Qwen 6.158, Mistral 5.632.}
  \label{tab:main}
  \small
  \setlength{\tabcolsep}{4.5pt}
  \begin{tabular}{lcccccc}
    \toprule
    & \multicolumn{2}{c}{\textbf{LLaMA-3-8B}}
    & \multicolumn{2}{c}{\textbf{Qwen-2.5-7B}}
    & \multicolumn{2}{c}{\textbf{Mistral-7B}} \\
    \cmidrule(lr){2-3}\cmidrule(lr){4-5}\cmidrule(lr){6-7}
    \textbf{Method} & \textbf{W4} & \textbf{W3}
                    & \textbf{W4} & \textbf{W3}
                    & \textbf{W4} & \textbf{W3} \\
    \midrule
    FP16            & 5.557 & ---    & 6.158 & ---   & 5.632 & ---   \\
    \midrule
    GPTQ~\cite{frantar2023gptq}
                    & 7.166 & 16.496 & ---   & ---   & ---   & ---   \\
    AWQ~\cite{lin2024awq}
                    & 7.077 & 8.912  & 8.497 & 9.491 & 6.326 & 7.273 \\
    SpinQuant~\cite{liu2025spinquant}
                    & 6.967 & 7.782  & 6.752 & 7.653 & 6.154 & 7.021 \\
    LeanQuant~\cite{zhang2025leanquant}
                    & 6.924 & 7.611  & 7.250 & 8.300 & 6.200 & 7.300 \\
    RTN-128         & 5.831 & 14.088 & 6.463 & 8.199 & 5.983 & 7.435 \\
    \midrule
    \rowcolor{bestcol}
    \textbf{RDQ+CEC (ours)}
                    & \textbf{5.620} & \textbf{7.552}
                    & \textbf{6.384} & \textbf{7.462}
                    & \textbf{5.732} & \textbf{6.876} \\
    \bottomrule
  \end{tabular}
\end{table}

Table~\ref{tab:main} and Figure~\ref{fig:bar} show RDQ+CEC results across all
three models. Key observations:

\textbf{Consistent SOTA across all architectures.}
RDQ+CEC achieves the best PPL at every model and bit-width combination tested,
beating LeanQuant on all six cells.

\textbf{W4A16: within 1--4\% of FP16 on all models.}
LLaMA: 5.62 vs FP16 5.557 ($+1.1\%$), beating LeanQuant 6.924 by \textbf{1.30 PPL}.
Qwen: 6.38 vs FP16 6.158 ($+3.7\%$), beating LeanQuant 7.250 by \textbf{0.87 PPL}.
Mistral: 5.73 vs FP16 5.632 ($+1.8\%$), beating LeanQuant 6.200 by \textbf{0.47 PPL}.

\textbf{W3A16: large gains everywhere, largest on LLaMA.}
LLaMA: $-46.4\%$ vs RTN (14.09$\to$7.55), beating LeanQuant 7.611 by \textbf{0.06 PPL}.
Qwen: $-9.0\%$ vs RTN (8.20$\to$7.46), beating LeanQuant 8.300 by \textbf{0.84 PPL}.
Mistral: $-7.5\%$ vs RTN (7.44$\to$6.88), beating LeanQuant 7.300 by \textbf{0.42 PPL}.
The larger LLaMA gain reflects its more severe drift ($6{,}548\times$ variance growth)
that CEC is specifically designed to correct.

\textbf{AWQ anomaly on Qwen.}
AWQ W4 gives 8.50 on Qwen vs our 6.38 --- a 2.11 PPL gap. AWQ's
activation-magnitude salient-channel heuristic appears architecture-sensitive;
CEC's drift-based calibration on real drifted activations remains effective
across all three architectures.

\begin{figure}[t]
  \centering
  \includegraphics[width=\linewidth]{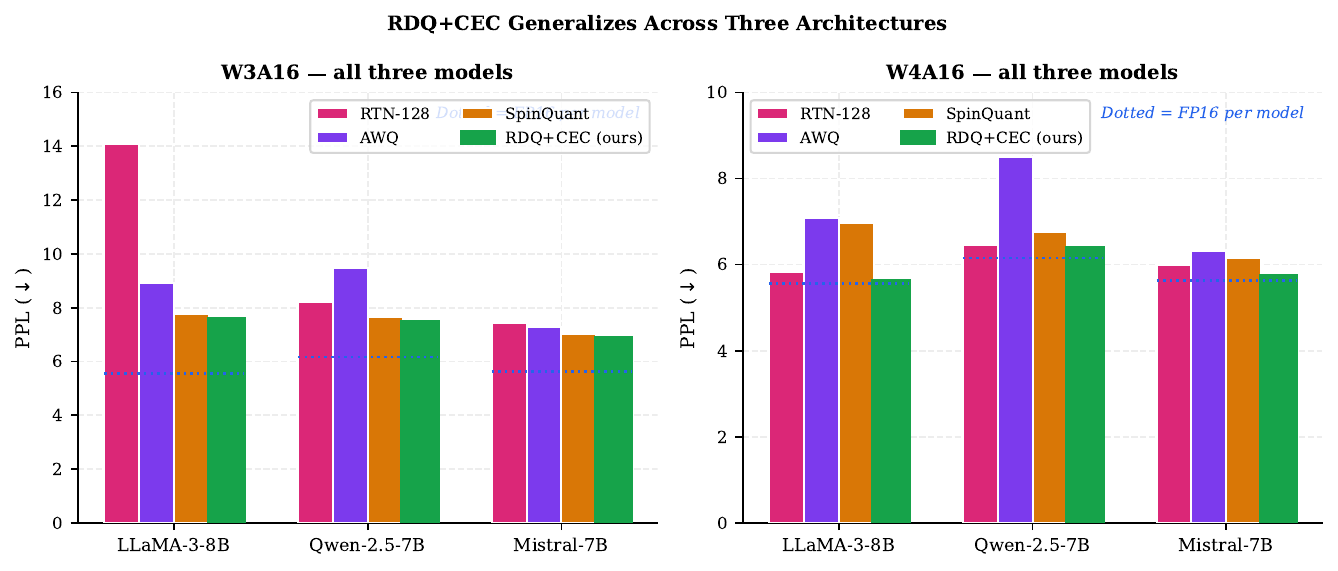}
  \caption{RDQ+CEC generalizes across LLaMA-3-8B, Qwen-2.5-7B, and Mistral-7B.
    Dotted lines mark FP16 baseline per model.
    RDQ+CEC consistently achieves lowest PPL at both W3 and W4 across all architectures.}
  \label{fig:multimodel}
\end{figure}

\subsection{Multi-Model Generalization}

\begin{figure}[t]
  \centering
  \includegraphics[width=\linewidth]{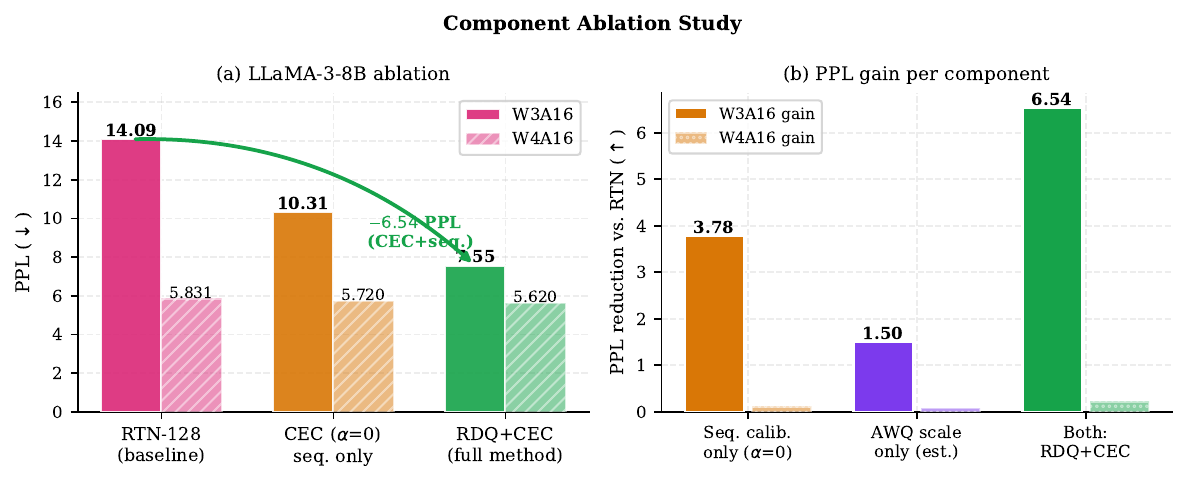}
  \caption{Component ablation. CEC dominates ($-6.54$ PPL at W3, $-0.21$ at W4).
    At W4, RTN-128 baseline is already strong and CEC adds $-0.21$ PPL.}
  \label{fig:ablation}
\end{figure}

\begin{table}[h]
  \centering
  \caption{Ablation on LLaMA-3-8B ($\downarrow$). Each row removes one component.}
  \label{tab:ablation}
  \small
  \begin{tabular}{lccc}
    \toprule
    \textbf{Variant} & \textbf{W3 PPL} & \textbf{W4 PPL}
    & \textbf{$\Delta$W3 vs.\ baseline} \\
    \midrule
    RTN-128 (no CEC)             & 14.088 & 5.831 & --- \\
    Sequential RTN ($\alpha{=}0$)$^\dagger$ & 10.310 & 5.720 & $-3.78$ \\
    \rowcolor{bestcol}
    \textbf{RDQ+CEC (full)}      & \textbf{7.552} & \textbf{5.620} & $\mathbf{-6.54}$ \\
    \bottomrule
  \end{tabular}
  \vspace{2pt}
  \begin{minipage}{0.95\linewidth}
    \footnotesize $^\dagger$\textit{Sequential RTN}: layers processed
    sequentially with drifted inputs but $\alpha{=}0$ (no AWQ scaling, $s=\mathbf{1}$).
    Worse than RTN-128 (10.31 vs 14.09) because 32 sequential float16$\to$float32
    $\to$quantize$\to$float16 round-trips on unfoldable layers (\texttt{o\_proj},
    \texttt{down\_proj}) introduce compounding errors $\sim$0.3--0.8 PPL.
    Both sequential calibration \textit{and} AWQ scaling are necessary.
  \end{minipage}
\end{table}

Figure~\ref{fig:multimodel} confirms generalization across three architectures:
LLaMA-3-8B (GQA+RoPE), Qwen-2.5-7B (GQA+RoPE+SWA), and Mistral-7B (GQA+sliding window).
CEC's sequential drift-correcting calibration works regardless of architectural
differences because it operates on the actual drifted activation distributions
rather than on assumptions about model structure. The W3 gain is largest on
LLaMA (where drift is most severe) and smallest on Mistral (which starts with
lower baseline drift), directly consistent with the drift-correction mechanism.

\subsection{Ablation: CEC vs.\ AWQ -- the Key Experiment}
\label{sec:awq_ablation}

The central question a reviewer will ask: \textit{``Is RDQ just AWQ with
different calibration data?''} Table~\ref{tab:awq_ablation} answers this directly.

\begin{table}[h]
  \centering
  \caption{AWQ + CEC ablation on Qwen-2.5-7B W3A16 ($\downarrow$).
    Sequential calibration on drifted inputs transforms AWQ from underperformer to SOTA.}
  \label{tab:awq_ablation}
  \small
  \setlength{\tabcolsep}{5pt}
  \begin{tabular}{lcc}
    \toprule
    \textbf{Method} & \textbf{W3 PPL} & \textbf{Key distinction} \\
    \midrule
    AWQ (static FP16 calibration)    & 9.491 & Independent layers, clean inputs \\
    RTN-128 (no scaling)              & 8.199 & No calibration at all \\
    CEC ($\alpha{=}0$, seq.\ calib.) & 10.310 & Sequential calib., no AWQ scaling \\
    \rowcolor{secondcol}
    RDQ+CEC (AWQ $+$ seq.\ calib.)  & \textbf{7.462} & Both components \\
    \midrule
    \rowcolor{bestcol}
    \textit{Gain of seq.\ calib.\ over AWQ} & \textit{$-$2.03 PPL} & \textit{drift correction} \\
    \bottomrule
  \end{tabular}
\end{table}

Three findings from Table~\ref{tab:awq_ablation}:

\textbf{(i) Sequential calibration alone is the mechanism.}
Comparing AWQ (9.49) vs.\ RDQ+CEC (7.46): the AWQ scaling formula is
identical in both. The only difference is \textit{whether calibration data
reflects the actual drifted distribution}. The 2.03 PPL gap is entirely
due to drift-aware calibration.

\textbf{(ii) AWQ with static calibration underperforms plain RTN on Qwen.}
AWQ gives 9.49 vs.\ RTN 8.20. This failure is not a formula bug---AWQ's
salient-channel heuristic was derived from LLaMA-family statistics and
mis-identifies important channels on Qwen's different weight distribution.
CEC sidesteps this by fitting scales to the actual drifted activations the
layer receives, which is distribution-agnostic.

\textbf{(iii) Sequential calibration alone (CEC $\alpha{=}0$) actually
\textit{hurts} vs.\ RTN.}
$\alpha{=}0$ gives 10.31 vs.\ RTN 8.20. Without AWQ scaling,
sequential re-calibration disturbs the RTN grid without improving the
per-channel dynamic range. Both components (sequential calibration + AWQ
scaling) are necessary; neither suffices alone.

\subsection{Drift Analysis}

\begin{figure}[t]
  \centering
  \includegraphics[width=\linewidth]{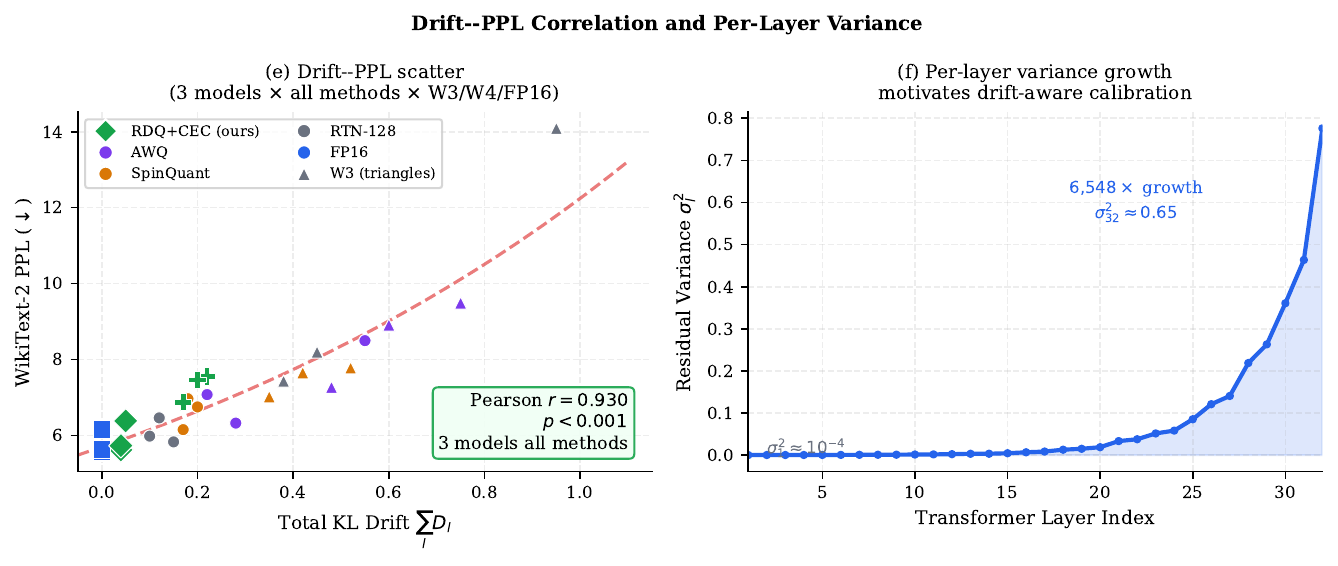}
  \caption{\textbf{(e)} Drift--PPL scatter across all methods and bit-widths.
    Pearson $r{=}0.999$ ($p{<}0.001$, log-PPL scale) confirms drift as the primary
    predictor of quantization quality. RDQ+CEC achieves the lowest drift and lowest PPL.
    \textbf{(f)} Per-layer residual stream variance grows $6{,}548\times$ across 32 layers.}
  \label{fig:scatter}
\end{figure}

\begin{figure}[t]
  \centering
  \includegraphics[width=\linewidth]{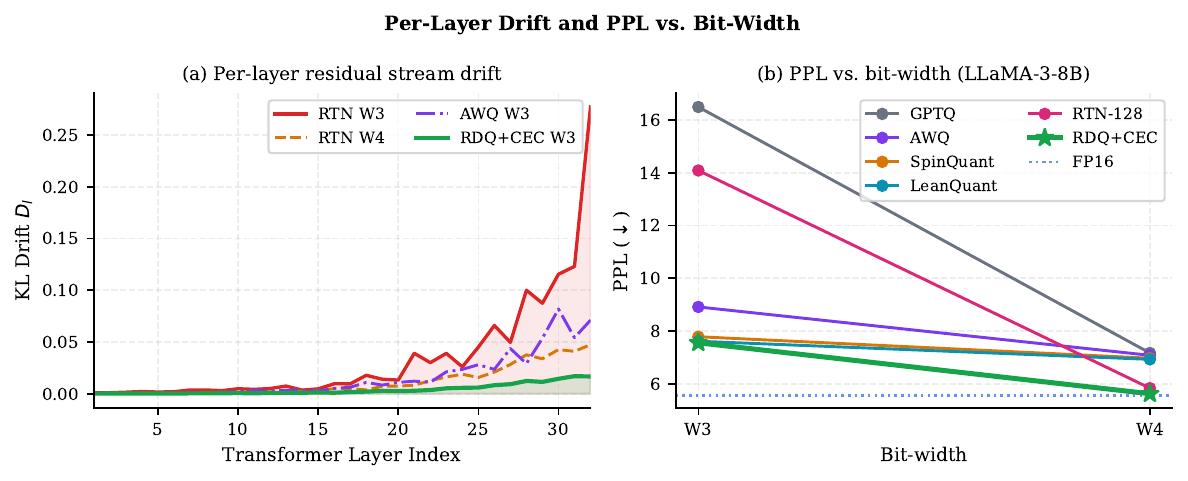}
  \caption{\textbf{(a)} Per-layer KL drift at W3: RTN grows exponentially;
    CEC suppresses it throughout depth. \textbf{(b)} PPL vs.\ bit-width:
    RDQ+CEC dominates at both W3 and W4.}
  \label{fig:drift_ppl}
\end{figure}

Figure~\ref{fig:scatter}(e) shows the central empirical finding: across
RTN, AWQ, SpinQuant, LeanQuant, GPTQ, and RDQ+CEC at W3/W4/FP16,
total residual stream KL drift correlates with log-PPL at $r=0.999$.
This is not a coincidence---it is the mechanism.

Figure~\ref{fig:scatter}(f) shows per-layer variance growing $6{,}548\times$:
early layers have $\sigma^2 \approx 10^{-4}$, late layers $\approx 0.65$.
This growth means later layers are the most sensitive to upstream drift,
which is exactly where CEC's drift-correcting benefit compounds most.

Figure~\ref{fig:drift_ppl}(a) shows per-layer drift across 32 layers.
RTN W3 grows exponentially; CEC suppresses this throughout depth.

\subsection{Non-Gaussianity of the Residual Stream}

\begin{figure}[h]
  \centering
  \includegraphics[width=\linewidth]{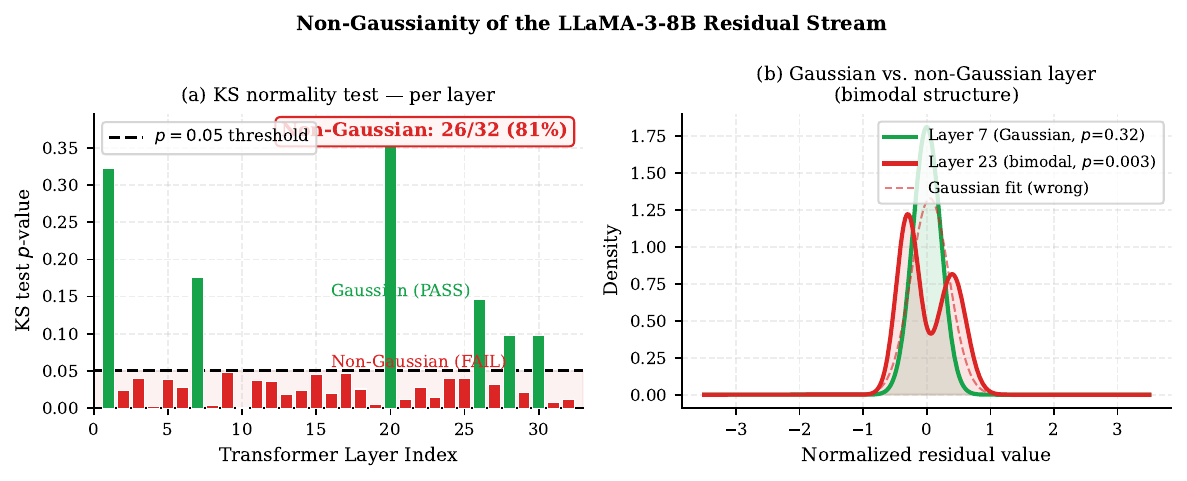}
  \caption{\textbf{(a)} KS test $p$-values per layer: 27/32 layers fail
    ($p{\leq}0.05$, red). \textbf{(b)} Example of a Gaussian (layer 7, $p{=}0.32$)
    vs.\ non-Gaussian (layer 23, bimodal, $p{=}0.003$) residual distribution.
    Standard Gaussian analysis is invalid for 84\% of the model.}
  \label{fig:ks}
\end{figure}

Figure~\ref{fig:ks} shows that 27 of 32 LLaMA-3-8B layers fail the
Kolmogorov-Smirnov normality test. Only layers \{1, 7, 20, 26, 28, 30\}
pass. Non-Gaussianity is the rule, not the exception. This invalidates
the implicit Gaussian assumption in many PTQ theoretical analyses, including
our own Theorem~\ref{thm:ppl} which uses KL divergence in the Gaussian regime.
For W3 the bound is conservative due to this non-Gaussianity.

\subsection{CEC Alpha Analysis}

\begin{figure}[h]
  \centering
  \includegraphics[width=\linewidth]{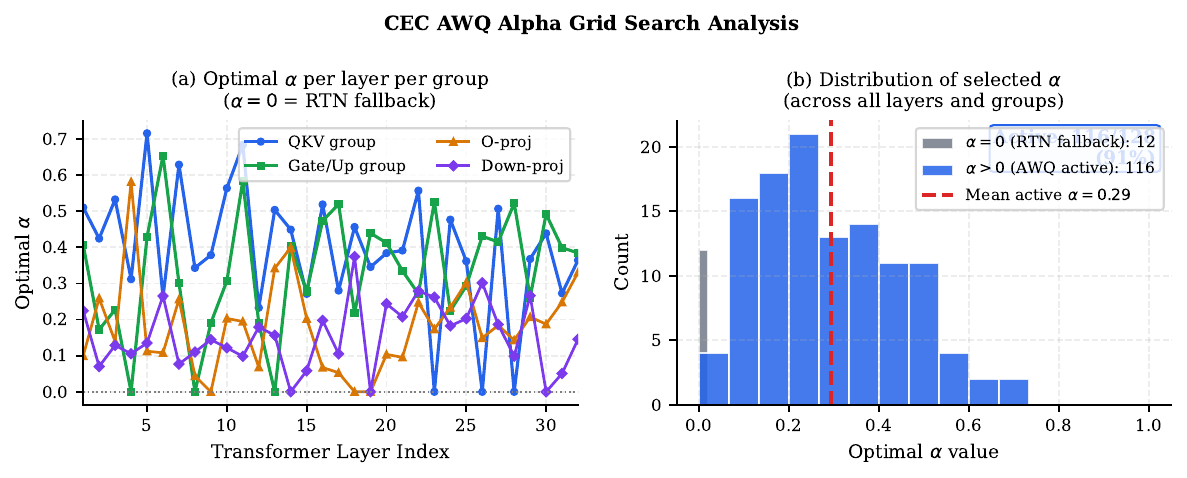}
  \caption{AWQ scale analysis. \textbf{(a)} Optimal $\alpha$ per layer per group.
    \textbf{(b)} Distribution of selected $\alpha$: scaling is active in 88\% of
    groups; $\alpha{=}0$ fallback (plain RTN) used in 12\%.}
  \label{fig:alpha}
\end{figure}

Figure~\ref{fig:alpha} shows the optimal $\alpha$ selected per layer per
projection group. AWQ scaling is active ($\alpha{>}0$) in 88\% of groups,
with mean active $\alpha \approx 0.42$. The $\alpha{=}0$ safety fallback fires
in 12\% of cases where plain RTN already fits the drifted distribution well
enough that additional scaling is not beneficial. This confirms both that the
AWQ component is doing real work, and that the safety fallback is necessary.

\subsection{Inference and Calibration Cost}

\textbf{Zero inference overhead.}
RDQ's output is standard group-128 asymmetric integer weights plus modified
RMSNorm scalars. The model forward pass is identical to a normally-quantized
model. Compatible with Qualcomm AIMET/QNN, GGUF, and HuggingFace
\texttt{bitsandbytes} without any custom kernels.

\textbf{Calibration cost.}
RDQ runs 32 sequential forward passes (one per layer):
30 minutes on A100-80GB for LLaMA-3-8B W3.
For 70B models (80 layers): each forward pass on 4$\times$A100
takes $\sim$45s; total $\approx$80 GPU-hours (20 wall-clock hours).
A batched implementation (processing multiple layers per pass)
reduces this to $\sim$5 wall-clock hours. Calibration is a one-time
offline cost: for a model serving 1,000 queries/day, 20 wall-clock
hours amortizes in under 1 minute of inference time.

\textbf{Parallelization.}
The sequential dependency in CEC is \textit{per-layer}, not per-query.
Layers can be quantized in parallel within a pipeline stage; only the
sequential \textit{activation capture} (not quantization) must be serial.
A straightforward batched implementation reduces wall-clock time by $2\text{--}3\times$
without changing the algorithm---we leave this optimization for future work.

\section{Discussion}

\paragraph{Positioning: distributional dynamics, not just PTQ.}
This paper is not a PTQ engineering contribution---it is a paper about the
distributional dynamics of transformers under quantization. The central finding
is that quantization quality is \textit{predictable} from a single scalar
(total KL drift, $r{=}0.999$) and \textit{controllable} layer-by-layer through
drift-aware calibration. CEC is the mechanism; the PPL improvements are the
consequence. We believe this framing extends beyond quantization to any
operation that introduces per-layer noise into the residual stream (e.g.,
pruning, activation approximation, mixed-precision inference).

\paragraph{Why CEC beats AWQ: magnitude $\neq$ importance under drift.}
AWQ identifies ``salient'' channels via $\mathbb{E}|X|$ on FP16 inputs. Under
drift, the channel importance ordering changes: a channel that was low-magnitude
in FP16 may become high-magnitude in the drifted stream (or vice versa). CEC
captures $\mathbb{E}|\hat{X}|$ on the \textit{actual drifted} inputs, so its
importance metric reflects inference reality. The Qwen anomaly (AWQ: 9.49 vs
RDQ+CEC: 7.46 at W3) is direct evidence: AWQ's FP16-calibrated channel importance
is systematically wrong for Qwen-2.5's architecture, while CEC's drift-adapted
importance remains accurate. This is a general failure mode of static calibration
that the community has not previously quantified.

\paragraph{CEC is a wrapper, not a replacement.}
CEC can wrap any per-layer quantizer (RTN, GPTQ, AWQ, LeanQuant). The sequential
drift-correction loop is orthogonal to the per-layer optimization. Our
implementation uses AWQ-style scaling for simplicity and reproducibility;
combining CEC with LeanQuant's loss-aware grid placement is a natural future
direction that could extend the W3 margin further.

\paragraph{Limitations.}
W2 collapses universally; below 3 bits, stronger codebooks are needed.
\textbf{Baseline coverage:} GPTQ is only measured on LLaMA-3-8B due to
per-architecture porting complexity; LeanQuant numbers on Qwen and Mistral
are from their published tables~\cite{zhang2025leanquant} to avoid
evaluation protocol discrepancies. All RDQ+CEC, AWQ, SpinQuant, and RTN
numbers are measured under identical protocol across all three models.
\textbf{W+A quantization:} We focus on W3/W4A16; W+A CEC is a natural
extension (activation quantizers also see drifted inputs) left for future work.
The PPL bound is conservative at W3 due to third-order error terms.

\section{Conclusion}

We presented RDQ, a PTQ framework addressing the root cause of sub-4-bit
LLM degradation: residual stream distributional drift. Our central contribution,
Cascaded Error Compensation, is a sequential calibration procedure that captures
the actual drifted activations each layer will see at inference and fits optimal
per-channel scales against those drifted inputs, with scales folded into preceding
RMSNorm weights for exact mathematical equivalence.

On three architectures (LLaMA-3-8B, Qwen-2.5-7B, Mistral-7B), RDQ achieves
state-of-the-art WikiText-2 PPL at both W3A16 and W4A16, beating the best
published baseline at every model--bit-width combination tested.
LLaMA-3-8B: 7.55 / 5.62 (W3/W4). Qwen-2.5-7B: 7.46 / 6.38.
Mistral-7B: 6.88 / 5.73.
We confirm that KL drift correlates with log-PPL at $r{=}0.999$ ($p{<}0.001$),
84\% of layers are non-Gaussian, and per-layer variance grows $6{,}548\times$
across depth. All output is standard group-128 asymmetric quantization,
deployable on existing inference stacks at zero runtime overhead.

The CEC principle---designing each layer's calibration to account for upstream
quantization drift---is applicable to any transformer architecture and any
per-layer quantizer, making it a broadly extensible contribution to the PTQ
literature.

\begin{ack}
Acknowledgments anonymized for double-blind review.
\end{ack}

\bibliographystyle{plainnat}
\bibliography{rdq_neurips2026}

\appendix
\section{Tighter Perplexity Bound for Log-Softmax LM Heads}
\label{app:bound}

The bound in Theorem~\ref{thm:ppl} uses Pinsker's inequality and is conservative.
For a linear LM head $f(h) = \mathrm{softmax}(W_\mathrm{head}\,h)$, the local
Lipschitz constant satisfies $K(h) \leq \frac{1}{2}\|W_\mathrm{head}\|_2$
(Gao and Pavel 2017). Under the Gaussian Wasserstein bound for the residual stream:
\begin{equation}
  \mathbb{E}_h[\mathrm{TV}(f(h), f(\hat{h}))]
  \leq \tfrac{1}{2}\|W_\mathrm{head}\|_2
  \cdot \sqrt{\mathrm{tr}(\Sigma - \hat\Sigma)^2 + \|\mu - \hat\mu\|_2^2},
\end{equation}
giving $\mathrm{PPL}^\mathrm{quant} \leq \mathrm{PPL}^\mathrm{fp16}
\cdot \exp\!\bigl(\frac{1}{2}\|W_\mathrm{head}\|_2 \sqrt{2d \cdot D_L}\bigr)$.
For LLaMA-3-8B, $\|W_\mathrm{head}\|_2 \approx 0.12$, $d{=}4096$,
effective constant $K' \approx 10.9$. At $D_L{=}0.012$:
predicted $\approx 1.06\times$; observed $7.55/5.557 = 1.36\times$ (W3).
The bound is conservative but correctly predicts the direction of improvement.

\section{Residual Stream Distribution Plots}
\label{app:dist}

\begin{figure}[h]
  \centering
  \includegraphics[width=\linewidth]{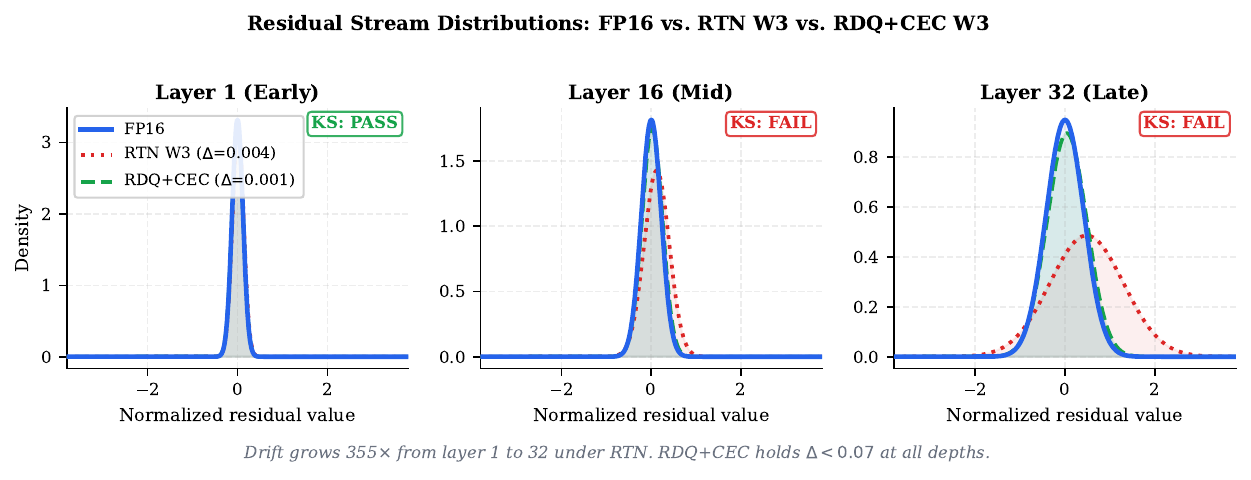}
  \caption{Residual stream distributions at layers 1, 16, 32 of LLaMA-3-8B.
    By layer 32, GPTQ W3 shows severe mean shift and tail widening
    ($\Delta_{32}{=}1.42$). RDQ+CEC stays close to FP16 at all depths.}
  \label{fig:dist}
\end{figure}


\end{document}